\documentclass[sigconf]{acmart}
\renewcommand\footnotetextcopyrightpermission[1]{} % removes footnote with conference information in first column
\usepackage{amsmath}  
\usepackage{multirow} 
\usepackage{makecell}
\usepackage{colortbl}

\usepackage{amssymb}
\definecolor{mygray}{gray}{0.94}
\usepackage{xcolor}
\definecolor{mylinkcolor}{HTML}{0000FF} % 蓝色
\definecolor{mycitecolor}{HTML}{008000} % 绿色
\definecolor{myurlcolor}{HTML}{FF0000} % 红色
\definecolor{ccr}{RGB}{255,0,150}  

\newcommand{\myurl}[2]{\textcolor{ccr}{\url{#1}}}
\hypersetup{%
    colorlinks = true,
    linkcolor = blue,
    citecolor = green,
    urlcolor = red
}
\begin{document}

\title{HoloTime: Taming Video Diffusion Models for Panoramic 4D Scene Generation}

\author{Haiyang Zhou}
\authornote{Both authors contributed equally to this research.}
\email{zhouhaiyang000@gmail.com}
\orcid{0009-0004-3616-9120}
\affiliation{%
  \institution{School of Electronic and Computer Engineering, Peking University}
  \city{Shenzhen}
  \country{China}
}

\author{Wangbo Yu}
\authornotemark[1]
\email{yuwangbo98@gmail.com}
\orcid{0000-0003-4387-8967}
\affiliation{%
  \institution{School of Electronic and Computer Engineering, Peking University}
  \institution{Peng Cheng Laboratory}
  \city{Shenzhen}
  \country{China}}

\author{Jiawen Guan}
\email{3291495912@qq.com}
\orcid{0009-0006-5148-1379}
\affiliation{%
  \institution{Harbin Institute of Technology, Shenzhen}
  \city{Shenzhen}
  \country{China}}

\author{Xinhua Cheng}
\email{chengxinhua@stu.pku.edu.cn}
\orcid{0000-0001-9034-279X}
\affiliation{%
  \institution{School of Electronic and Computer Engineering, Peking University}
  \city{Shenzhen}
  \country{China}}

\author{Yonghong Tian}
\authornote{Corresponding authors.}
\email{yhtian@pku.edu.cn}
\orcid{0000-0002-7427-8764}
\affiliation{%
  \institution{School of Electronic and Computer Engineering, Peking University}
  \institution{Peng Cheng Laboratory}
  \city{Shenzhen}
  \country{China}}

\author{Li Yuan}
\authornotemark[2]
\email{yuanli-ece@pku.edu.cn}
\orcid{0000-0002-2120-5588}
\affiliation{%
  \institution{School of Electronic and Computer Engineering, Peking University}
  \institution{Peng Cheng Laboratory}
  \city{Shenzhen}
  \country{China}}

%\renewcommand{\shortauthors}{Zhou et al.}

\begin{abstract}
The rapid advancement of diffusion models holds the promise of revolutionizing the application of VR and AR technologies, which typically require scene-level 4D assets for user experience. 
Nonetheless, existing diffusion models predominantly concentrate on modeling static 3D scenes or object-level dynamics, constraining their capacity to provide truly immersive experiences. % background
To address this issue, we propose \textit{HoloTime}, a framework that integrates video diffusion models to generate panoramic videos from a single prompt or reference image, along with a 360-degree 4D scene reconstruction method that seamlessly transforms the generated panoramic video into 4D assets, enabling a fully immersive 4D experience for users.
Specifically, to tame video diffusion models for generating high-fidelity panoramic videos, we introduce the \textit{360World dataset}, the first comprehensive collection of panoramic videos suitable for downstream 4D scene reconstruction tasks.
With this curated dataset, we propose \textit{Panoramic Animator}, a two-stage image-to-video diffusion model that can convert panoramic images into high-quality panoramic videos. Following this, we present \textit{Panoramic Space-Time Reconstruction}, which leverages a space-time depth estimation method to transform the generated panoramic videos into 4D point clouds, enabling the optimization of a holistic 4D Gaussian Splatting representation to reconstruct spatially and temporally consistent 4D scenes.
To validate the efficacy of our method, we conducted a comparative analysis with existing approaches, revealing its superiority in both panoramic video generation and 4D scene reconstruction. This demonstrates our method's capability to create more engaging and realistic immersive environments, thereby enhancing user experiences in VR and AR applications. More results are available on the project page: \myurl{https://zhouhyocean.github.io/holotime/}{https://zhouhyocean.github.io/holotime/}.
\end{abstract}


\begin{teaserfigure}
  \includegraphics[width=\textwidth]{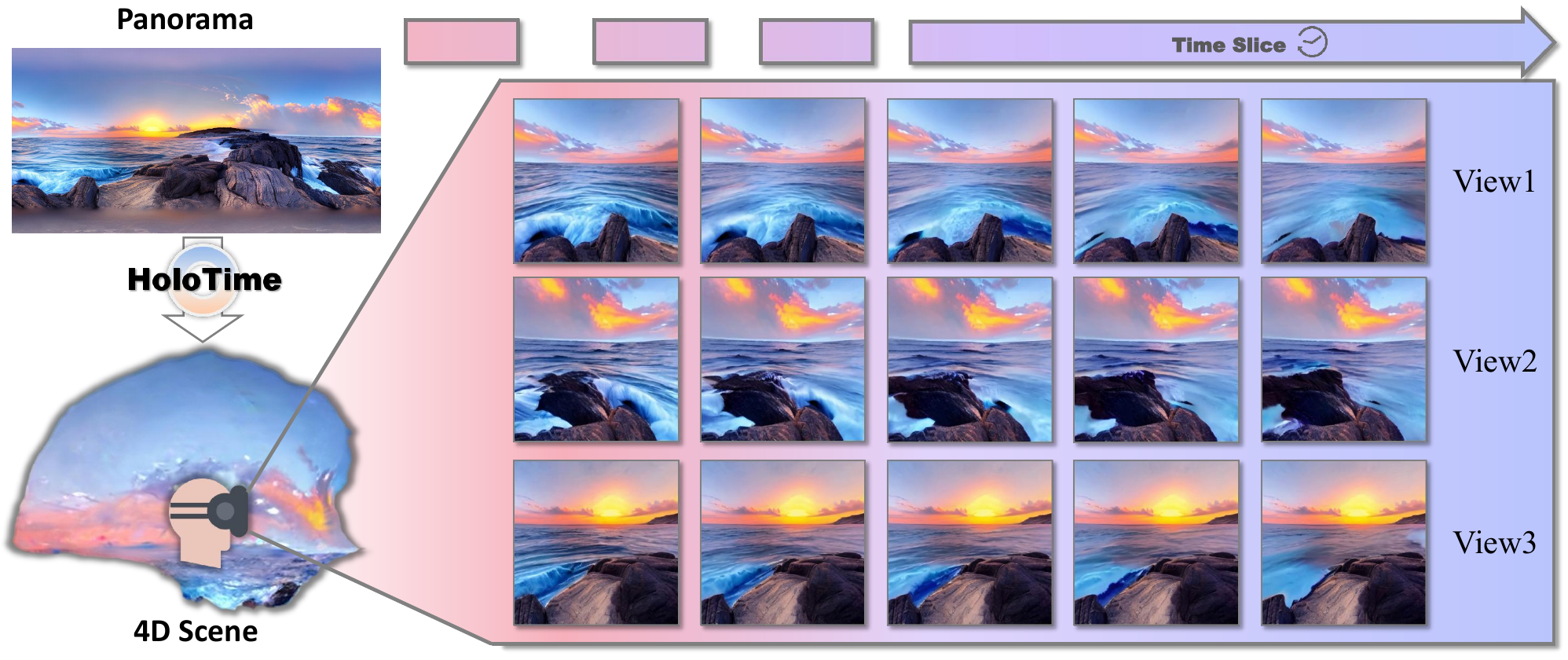}
  \caption{HoloTime accepts either a user-provided or model-generated panoramic image as input, and transforms it into an immersive 360-degree 4D scene, enabling a virtual roaming experience.}
  \Description{}
  \label{fig:teaser}
\end{teaserfigure}

\maketitle

\section{Introduction}
The advancement of diffusion models for text-to-image generation~\cite{DDPM,LDM} has sparked a burgeoning interest in video and 3D content creation, leading to significant milestones. Building upon the foundation of 3D content generation, 4D content generation—which integrates the 
temporal dimension into 3D data—holds immense potential for applications and promises to revolutionize the fields of Augmented Reality (AR) and Virtual Reality (VR). 
% Withthe evolution of generative models, Artificial Intelligence-Generated Content (AIGC) has become a highly prospective domain of research. The diffusion models for text-to-image have significantly transformed the field of computer vision and are now extensively utilized across a myriad of tasks and scenarios. Following the breakthroughs in text-to-image, there has been a surge of interest in video and 3D content generation, which have achieved considerable milestones. Building on the foundation of 3D content generation, 4D content generation, which incorporates the temporal dimension into 3D data, presents an expansive potential for applications and could instigate novel revolutions in the realms of Augmented Reality (AR) and Virtual Reality (VR). 
Scene-level 4D content generation has the potential to provide users with a more liberating and immersive experience, allowing for greater freedom in virtual navigation. However, unlike the advancements in video and 3D content generation, 4D content generation has lagged behind due to the scarcity of high-quality annotated data, particularly for large-scale 4D scenes. Consequently, current 4D generation methods are primarily limited to generating object-level dynamics~\cite{4dfy,consistent4d,dreamgaussian4d,alignyourgaussians,4dgen} or creating forward-facing scenes with restricted viewing freedom~\cite{cat4d}.
%VividDream和4K4D放到related work讨论
% VividDream~\cite{} employs outpainting techniques to expand the boundaries of a scene and generates a series of videos based on diverse camera trajectories for the purpose of scene reconstruction. The generation of local dynamics can lead to issues with global consistency and the emergence of artifacts. 

% The generation of large-scale 4D scenes could provide users with a more liberating and immersive experience, facilitating a higher degree of freedom in virtual navigation. However, in contrast to advancements in video and 3D content generation, 4D content generation has lagged due to the scarcity of high-quality annotated data, especially for extensive 4D scenes. The existing 4D generation methods are typically based on the priors of video diffusion models and multi-view diffusion models, applicable only to objects, and while some have attempted to generate 4D scenes from a single image, the limited views result in diminutive scenes with restricted freedom of experience. VividDream~\cite{} employs outpainting techniques to expand the boundaries of a scene and generates a series of videos based on diverse camera trajectories for the purpose of scene reconstruction. The generation of local dynamics can lead to issues with global consistency and the emergence of artifacts. 
To cultivate a 360-degree immersive experience, an effective strategy is to harness the 360° field of view provided by panoramic images. While established methods~\cite{dreamscene360,holodreamer,fastscene} already utilize panoramic images for 3D generation, introducing dynamics to these images for 4D generation is not straightforward due to the lack of large-scale panoramic video training data.
% Due to differences in Field of View (FoV) and projection methods, panoramic images exhibit significantly distinct data distributions compared to conventional perspective images. They are characterized by two notable features that set them apart from typical images: 1. Due to the Equirectangular Projection (ERP), panoramic images can experience distortion. The distortion becomes more severe towards the polar regions (top and bottom), often leading to curved shapes and motion trajectories.
% 2. With a horizontal field of view of 360 degrees, the leftmost and rightmost parts of a panoramic image are continuous, creating a cyclic effect. Because of these attributes, conventional video generation models are not equipped to produce panoramic videos effectively. 4k4dgen, which is based on a perspective video diffusion model, first introduces panoramic videos generation into the 4D scene generation domain. Its Panoramic Denoiser address the challenges of noise reduction in panoramic videos. However, denoising used local small windows may constrain the range of motion and lead to inconsistencies across views. In contrast, our method, introduces the first dataset for panoramic video and panoramic 4D scene generation, effectively addressing the gap in panoramic video datasets. Furthurmore, we propose a set of carefully designed techniques to transform general Image-to-Video (I2V) models into models capable of producing panoramic video content with high fidelity.
To address these challenges, we introduce \textit{HoloTime}, a 360-degree 4D scene generation framework. It accepts a user-provided or model-generated~\cite{flux} panoramic image as input, and leverage a \textit{Panoramic Animator} to transform the panoramic image into a panoramic video featuring realistic visual dynamics and a fixed camera pose. Following this, we present a \textit{Panoramic Space-Time Reconstruction} technique that reconstructs a high-fidelity 4D scene based on the generated video, thereby enabling an immersive user experience.
To build the framework, we first curated a large-scale dataset of fixed-camera panoramic videos, accompanied by corresponding prompt descriptions, which we term the \textit{360World dataset}. We then train the \textit{Panoramic Animator} on this dataset to convert panoramic images into panoramic videos, employing a two-stage motion-guided generation process to enhance its performance.
After generating the panoramic video, we utilize the \textit{Panoramic Space-Time Reconstruction} technique for high-fidelity 4D scene reconstruction. Specifically, we begin by employing a panoramic optical flow estimation model~\cite{panoflow} in conjunction with a narrow field-of-view depth estimation model~\cite{marigold} to perform space-time depth estimation, ensuring depth consistency across both temporal and spatial dimensions. This aligned depth is used to convert the panoramic video into dynamic point clouds, optimizing a spatially and temporally consistent 4D-GS representation, which ultimately provides an immersive experience in virtual and augmented reality environments.

Our contributions are encapsulated as follows:
\begin{itemize}
\item We present \textit{Panoramic Animator}, accompanied by a two-stage motion-guided generation strategy that seamlessly transforms panoramic images into dynamic panoramic videos, all while preserving the spatial characteristics of the original images, facilitating downstream 4D reconstruction tasks.
% which leverages existing I2V models for hybrid data fine-tuning and employs a two-stage motion-guided generation approach that is congruent with the spatial characteristics of panoramic images, while also exhibiting pronounced motion and generalization capabilities.
\item We introduce \textit{Panoramic Space-Time Reconstruction}, which employs cutting-edge techniques to enable the temporal and spatial alignment of depth estimation for panoramic videos, enabling a holistic 4D scene reconstruction through the 4D-GS representations.
% We incorporate \textit{Panoramic Space-Time Reconstruction}, which, with the assistance of panoramic optical flow estimation and the expertise from perspective view depth estimation methods, enables the temporal and spatial alignment of depth estimation for panoramic videos, enabling panoramic 4D reconstruction through the use of 4D representations.
\item We contribute the \textit{360World dataset}, the first comprehensive collection of panoramic videos captured with fixed cameras. 
We believe this dataset can not only helps address gaps in 360-degree 4D scene generation but also holds promise for advancing future 4D generation efforts.
\end{itemize}

\section{Related Works}
% Head 2
\subsection{Diffusion-based Image and Video Generation.}

As supported by previous works \cite{DDPM,LDM,Imagen,DALLE2,GLIDE}, diffusion models have markedly enhanced the capacity to generate 2D images. 
%Particularly, Stable Diffusion (Rombach et al., 2022) has refined the operation of diffusion models (DMs) within the latent space of autoencoders, achieving a commendable equilibrium between computational efficiency and superior image quality.
The application of diffusion models also extends to the domain of video generation. VDM \cite{VDM} first introduced the space-time factorized U-Net architecture, extending the original 2D U-Net for video modeling and generation. Subsequently, many methods\cite{imagenvideo,makeavideo,magicvideo,modelscope,animatediff,alignyourlatent} have followed this similar architecture, achieving significant progress in generating high-quality video content from text. Some models \cite{SVD,i2vgen,dynamicrafter} utilize image control for video generation to achieve a higher degree of customization and consistency in the content produced. Recent studies \cite{videopoet,latte,cogvideox} have employed the DiT (Diffusion Transformer) architecture for video generation, also achieving impressive performance. Although current video generation is largely confined to perspective videos, 360DVD \cite{360dvd} has conducted explorations to panoramic video text-driven generation through training an adapter. Furthermore, Imagine360 \cite{imagine360} is capable of lifting input perspective videos to panoramic videos.

\subsection{3D Scene Generation.}
The emergence of NeRF \cite{nerf} and 3D Gaussian Splatting (3D-GS) \cite{3d-gs} has propelled significant advancements in the field of 3D reconstruction \cite{mipnerf,blocknerf,aenerf,nofa,evagaussians,neuralgs}. Additionally, there is growing exploration into leveraging 2D diffusion models to facilitate 3D generation. Early works \cite{magic3d, prolificdreamer, dreamgaussian,hifi123,progressive3d,repaint123,magic123} employ the Score Distillation Sampling (SDS) \cite{dreamfusion} to distill 3D representations from 2D diffusion models. However, most of these efforts focus on object-level generation, and multiple iterations of distillation could lead to excessive saturation.
% Magic3D [11], ProlificDreamer [12], HiFi-123 [32], Progressive3D [33] and DreamGaussian
%The development of 3D scene generation technology has been closely tied to the evolution of 3D representation methods. Leading to the birth of numerous 3D-centric object generation methods based on 3D-GS. %Multi-view models and video models possess strong 3D perception capabilities, enabling the generation of more 3D-consistent views, which can then be used for 3D reconstruction. 
Some works \cite{luciddreamer,text2immersion,wonderjourney,wonderworld} are based on outpainting to generate larger-scale scenes and conduct lifting from 2D to 3D using depth estimation models. 
%Methods such as LucidDreamer, Text2Immersion, RealmDreamer, and WonderJourney have enabled the generation of more extensive 3D scenes. 
To achieve more immersive virtual roaming, many methods \cite{dreamscene360,holodreamer,fastscene,layerpano} choose panoramic images with a spherical field of view (FoV) as the scene representation. 
However, these methods can only create static 3D scenes without any motion, restricting their vibrancy.

\begin{figure*}
  \includegraphics[width=\textwidth]{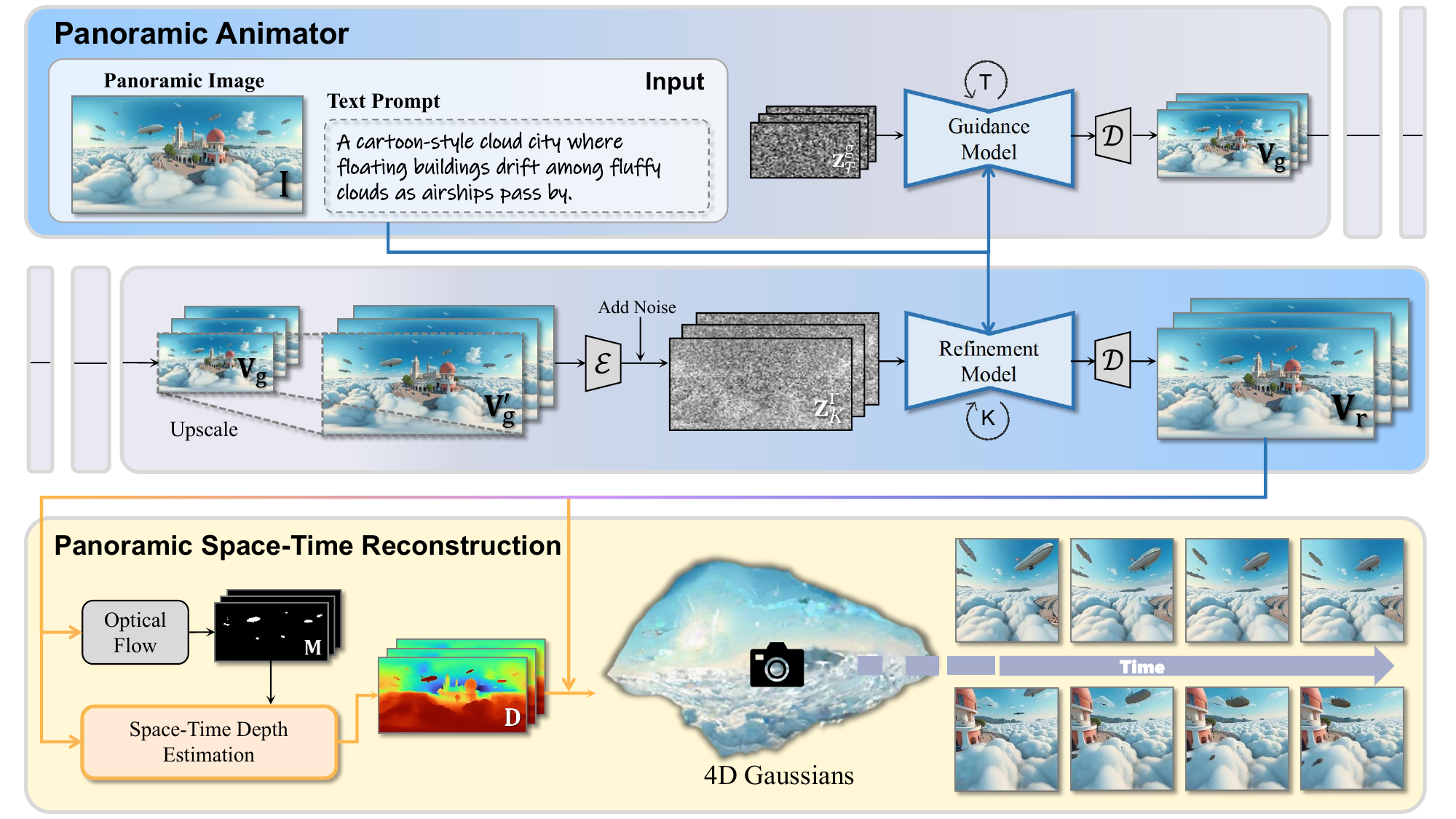}
  \caption{Overview of HoloTime. Beginning with a user-provided or model-generated panoramic image as input, we first use the Panoramic Animator to generate panoramic videos in two stages. The guidance model generates a coarse video in the first stage, which is then refined by the refinement model in the second stage, creating the final panoramic video for 4D reconstruction. Subsequently, we perform Panoramic Space-Time reconstruction to lift the panoramic video to a 4D scene. We employ optical flow for space-time depth estimation to achieve spatial and temporal alignment, thus obtaining a 4D initialized point cloud. Finally, we employ a 4D-GS method for the final scene reconstruction representation. }
  \label{fig:method}
\end{figure*}

\subsection{4D Scene Generation.}
%4dgen consistent4d dreamgaussian4d animate124 align your gaussian 4dfy
The achievements in video generation and 3D generation have indeed spurred the exploration of 4D generation tasks. Many object-level generation works \cite{4dfy,consistent4d,dreamgaussian4d,alignyourgaussians,4dgen} leverage video diffusion models and multi-view diffusion models \cite{zero123,mvdream,syncdreamer} that have 3D aware ability for optimization or reconstruction of 4D objects. CAT4D \cite{cat4d} introduces a more unified multi-view video diffusion model for limited-scale 4D scene generation. VividDream \cite{vividdream} employs an outpainting method to expand the scene scale, but it suffers from artifacts. For the generation of panoramic 4D scenes, 4K4DGen \cite{4k4dgen} proposes a novel denoising method to adapt the perspective video diffusion model for panoramic video generation. However, the denoising of multiple small fixed window leads to restricted motion ranges. DynamicScaler \cite{dynamicscaler} improves this by continuously shifting the angles of the windows during the denoising process. In contrast, our method directly animate the entire panoramic image, allowing for a more significant range of motion and better global consistency in the panoramic space.

\section{Methodology}

This section begins with a concise overview of the Diffusion Model in Sec. \ref{sec:pre}. Subsequently, we introduce  \textit{Panoramic Animator} and its techniques in Sec. \ref{sec:animator} and delve into the process of conducting \textit{Panoramic Space-Time Reconstruction} in Sec. \ref{sec:reconstruction}. Ultimately, the explanation of \textit{360World dataset} is presented in Sec. \ref{sec:dataset}. The overall framework of our method is illustrated in Fig. \ref{fig:method}.

%Given a static panoramic image as input, our method generates a large-scale 4D dynamic scene, allowing for a wide range of camera movements and a high degree of exploration freedom. We first introduce our proposed panoramic video dataset 360World in Sec.3.2, which will provide data support for our panoramic video generation. In the animation phase, Panoramic Animator will generate panoramic video from input panorama image.  will be elaborated in Section 3.3. Finally, we introduce Panoramic Space-Time reconstruction in Sec.3.4. We align the depth of panoramic video in space and time through Spacetime Depth Estimation, and leverage 4D representation methods to achieve 4D reconstruction.

\subsection{Preliminaries}\label{sec:pre}
\subsubsection{Diffusion Model}
Diffusion models \citep{DDPM} typically consist of two processes: forward diffusion process $q$ and reverse denoising process $p_\theta$. Given an input signal $\mathbf{x}_0\sim(q_0, \mathbf{x}_0)$, 
the forward process gradually introduces noise to $\mathbf{x}_0$. As the time steps increase, the level of noise gradually intensifies, ultimately resembling a Gaussian noise. This could be defined as follows: 
\begin{equation}
\mathbf{x}_t=\mathcal{Q}(\mathbf{x}_0, t)=\alpha_t \mathbf{x}_0 + \sigma_t \epsilon
\label{eq:addnoise}
\end{equation} 
where $t = 1,\dots,T$ and $\epsilon\sim\mathcal{N}(0, \mathbf{I})$. The hyper-parameters $\alpha_t$ and $\sigma_t$ satisfy the equation $\alpha^2_t+\sigma^2_t=1$. The reverse denoising process $p_\theta$ aims to optimize a noise predictor $\sigma_\theta$ to effectively remove noise, using the following loss function: 
\begin{equation}
\mathcal{L}=\mathbb{E}_{t\sim\mathcal{U}(0,1), \epsilon \sim \mathcal{N}(0, \mathbf{I})}\left[\left\|\epsilon-\epsilon_\theta\left(\mathbf{x}_t, t\right)\right\|_2^2\right]
\label{eq:diffusion_loss}
\end{equation}
Latent Diffusion Models \cite{LDM} enhance computational efficiency by employing a pre-trained VAE. Initially, the clean data $x_0$ is encoded into latent code using the VAE encoder $\mathcal{E}$. The diffusion and denoising processes are then performed in the latent space. Finally, the denoised latent code is decoded back to the original space using the VAE decoder $\mathcal{D}$. In video diffusion models, the parameter $\epsilon_\theta$ typically employs either a U-Net architecture \cite{magicvideo, makeavideo,animatediff,SVD} or a Transformer \cite{videopoet, latte, cogvideox}. Considering that most open-source Image-to-Video (I2V) models currently use the U-Net architecture, we adopt this architecture in our study. We build our panoramic video generation method based on DynamiCrafter \cite{dynamicrafter}, which effectively produces rich dynamic effects from input images.

\subsection{Panoramic Animator}\label{sec:animator}

Leveraging advanced I2V models, , we propose \textit{Panoramic Animator}, a novel method composed of three proposed mechanisms for generating panoramic videos from panoramic images.  %we introduce multiple mechanisms to retain the model’s pre-trained ability of generation while acquiring the capability to generate panoramic videos. 
%we introduce multiple mechanisms to acquire the capability to generate panoramic vi’‘deos while ensuring robust performance.
%regardless of whether the input images exhibit distinct panoramic geometric features.
we introduce Hybrid Data Fine-tuning (HDF) in Sec. \ref{sec:hybrid} and Two-stage Motion Guided Generation (MGG) in Sec. \ref{sec:guidance} to improve the model's performance, along with Panoramic Circular Techniques (PCT) in Sec. \ref{sec:circular} to enhance the visual effects of panoramic videos. 

\subsubsection{Hybrid Data Fine-tuning}\label{sec:hybrid}
Due to the significant distribution differences between general and panoramic videos, we avoid directly fine-tuning with panoramic video data to preserve the effective temporal priors of the pretrained video model. We incorporate additional video data for hybrid data fine-tuning. Landscape time-lapse videos exhibit significant motion. While typically recorded with perspective cameras, they share semantic and temporal similarities with panoramic videos, effectively bridging the gap between panoramic and standard videos in spatial-temporal data distribution and improving the model's generalization capabilities. We choose the ChronoMagic-Pro dataset \cite{chronomagic} for supplementation of training data. By searching for the keyword ``landscape'' in the annotated text, we filtered out irrelevant videos and collected $4,455$ relevant text-to-video pairs. These were then randomly mixed with our \textit{360World dataset} to create a hybrid dataset.

\subsubsection{Two-Stage Motion Guided Generation}\label{sec:guidance}
Panoramic videos offer a spherical viewing angle that contains a wealth of spatial information, typically showcasing localized fine motion rather than global large-scale motion. When training models with the same architecture and training data at varying resolutions, we find that the model prioritizes learning temporal information at lower resolutions and spatial information at higher resolutions. Therefore, we propose a two-stage motion-guided generation method: firstly, we generate a low-resolution coarse video that offers global motion guidance, and then generate refinement video with higher resolution.

%Existing video generation models are typically structured around 2D diffusion models and exhibit a spatial-temporal separation architecture. 
We fine-tune the guidance model $\mathcal{M}_{\text{g}}$ and refinement model $\mathcal{M}_{\text{r}}$ based on the pre-trained DynamiCrafter. During training, we fine-tune the spatial layers of the guidance model $\mathcal{M}_{\text{g}}$ using \textit{360World dataset} solely at low resolution, while keeping the temporal layers frozen. Subsequently, we fine-tune all the layers of the refinement model $\mathcal{M}_{\text{r}}$ using hybrid dataset at high resolution. In the inference phase, given the input panoramic image $\mathbf{I}$ and other optional conditions, like prompt, the guidance model $\mathcal{M}_{\text{g}}$ first generate a coarse video $\mathbf{V}_{\text{g}}$ from random latent Gaussian noise $\mathbf{z}^{\text{g}}_{T}$. This video $\mathbf{V}_{\text{g}}$ exhibits significant global motion aligned with the panoramic spatial features. We upscale $\mathbf{V}_{\text{g}}$ to high resolution $\mathbf{V}^{'}_{\text{g}}$ using a super-resolution model. Subsequently, We encode the video $\mathbf{V}^{'}_{\text{g}}$ into latent code and add noise to $K$ $(K<T)$ time steps. This could be represented by the following formula:

\begin{equation}
\mathbf{z}^{\text{r}}_{K}=\mathcal{Q}(\mathcal{E}(\mathbf{V}^{'}_{\text{g}}),K)
 \label{eq:encode_addnoise}
\end{equation}

We utilize the refinement model $\mathcal{M}_{\text{r}}$ for secondary denoising process, using user input as conditional control to enhance local motion details, resulting in the final video $\mathbf{V}_{\text{r}}$. This ensures that the generated video exhibits a strong dynamic effect on a global scale while also handling local details well.

%\begin{equation}
%\mathcal{N}(\text{z}_{0}, t)=\sqrt{\bar{\alpha}_{t}}\text{z}_{0}+\sqrt{1-\bar{\alpha}_{t}}\epsilon,\quad\alpha_{t}=1-\beta_{t},\quad\bar{\alpha}_{t}=\prod_{t=1}^{T}\alpha_{t}
% \label{eq:count}
%\end{equation

\subsubsection{Panoramic Circular Techniques}\label{sec:circular}
Horizontal end continuity is crucial for panoramic videos, influencing the user's seamless experience in subsequent 4D reconstructions. To achieve this, we create repeated sections at the left and right ends of the panoramic video and conduct blending after each denoising step throughout the generation process for complete continuity. Specifically, during inference process, A section from the left end of the reference image $\mathbf{I}$ is first copied and concatenated to the right end. After each denoising step, the left part of the corresponding latent code is blended into the right part, and thereafter, the right part is blended into the left part, continuing this alternating mixing. Following 360DVD \cite{360dvd}, We also modify the padding operation of the convolutional layers during denoising process of refinement model $\mathcal{M}_{\text{r}}$ by applying code at suitable positions to pad the left and right boundaries, ensuring pixel-level continuity. After generating the final video, the repeated part is cropped to yield the final seamless continuous panoramic video. 

\subsection{Panoramic Space-Time Reconstruction}\label{sec:reconstruction}

In the 4D lifting phase, we first introduce Space Aligned Depth Estimation for single panoramic image in Sec. \ref{sec:space_depth}, and then extend it to Space-Time Depth Estimation for panoramic video in Sec. \ref{sec:spacetime_depth}. Finally, we employ a 4D-GS representation method to complete the final 4D scene reconstruction in Sec. \ref{sec:representaion}.

\subsubsection{Space Aligned Depth Estimation}\label{sec:space_depth}
%Existing monocular depth estimation models are trained using perspective images or videos and cannot be directly applied to depth estimation of panoramic tasks. 
To estimate the depth of a single panoramic image using pre-trained perspective depth estimation models, 360MonoDepth\cite{360monodepth} proposes a general spatial alignment-based method that projects the panoramic image into multiple perspective images for individual depth estimation. The resulting depth maps are aligned and back-projected to form the panoramic depth map. Building on this concept, we define N groups of outward perspective view directions in the spherical panoramic field of view (FoV), denoted as $\{\mathbf{v}_1, \mathbf{v}_2, ..., \mathbf{v}_N\}$, where $\mathbf{v}_n\in \mathbb{R}^{h \times w \times 3}$. Each panoramic frame $\mathbf{f}^l\in \mathbb{R}^{H \times W \times 3}$ $(l=1,2,\ldots,L)$ will be reprojected into N perspective images %$\{\mathbf{I}^l_1, \mathbf{I}^l_2, ..., \mathbf{I}^l_N\}$
corresponding to these view directions according to the equirectangular projection. Subsequently, a perspective depth estimation model is used to estimate the depth of the perspective images, yielding N depth maps, denoted as $\{\mathbf{d}^l_1, \mathbf{d}^l_2, ..., \mathbf{d}^l_N\}$. Each depth map $\mathbf{d}^l_n\in \mathbb{R}^{h \times w \times 3}$ is assigned a learnable scale factor $\alpha^l_n\in \mathbb{R}$ and a learnable shift factor $\boldsymbol{\beta}^l_n\in \mathbb{R}^{h \times w}$ to adjust the depth values. Inspired by DreamScene360 \cite{dreamscene360}, we employ a learnable MLPs with parameters $\boldsymbol{\Theta}^{l}: \mathbb{R}^{3} \rightarrow \mathbb{R}$ to model a static geometric field of the panoramic frame $\mathbf{f}^l$. The input to the geometric field is the view directions $\mathbf{v}$, while the output is the depth values along the corresponding ray directions from the origin. We define the optimization goal for space alignment as follows:
\begin{align} \label{eq:spatial-align}
\min \operatornamewithlimits{\mathbb{E}}_{i \in S} \bigg\{&\lambda_{\text{depth}} \mathcal{L}_{\text{depth}}  + 
\lambda_{\text{scale}}\mathcal{L}_{\text{scale}} +
\lambda_{\text{shift}}\mathcal{L}_{\text{shift}}\bigg\} \\ \nonumber
\text{where} \quad &\mathcal{L}_{\text{depth}} = ||\operatorname{softplus}(\alpha^l_i) \mathbf{d}^l_i +  \boldsymbol{\beta}^l_i  - \operatorname{MLPs}(\mathbf{v}_i; \boldsymbol{\Theta}^{l}) || \\ \nonumber
&\mathcal{L}_{\text{scale}} = \|\operatorname{softplus}(\alpha_i^l)-1\|^2 \\ \nonumber
&\mathcal{L}_{\text{shift}} = \sum_{j,k} \left( (\boldsymbol{\beta}^l_{i,j,k+1} - \boldsymbol{\beta}^l_{i,j,k})^2 + (\boldsymbol{\beta}^l_{i,j+1,k} - \boldsymbol{\beta}^l_{i,j,k})^2 \right) \nonumber
\end{align}
where \( S \) denotes the index set for the perspective images involved in the current optimization. We initiate by estimating the depth map of the first frame $\mathbf{f}^1$, which is also the input reference image $\mathbf{I}$, based on Eq. \ref{eq:spatial-align} with $l=1$, involving all N perspective images. After optimization, the depth map $\mathbf{D}^1 \in \mathbb{R}^{H \times W}$ is determined using the geometric field $\boldsymbol{\Theta}^{1}$ with the panoramic direction $\mathbf{v}_{\text{pano}} \in \mathbb{R}^{H \times W \times 3}$ as input, computed as $\mathbf{D}^{l} = \text{MLPs}(\mathbf{v}_{\text{pano}}; \boldsymbol{\Theta}^{l}) $.

\subsubsection{Space-Time Depth Estimation}\label{sec:spacetime_depth}
Depth estimation for panoramic videos requires not only spatial alignment within each individual frame but also temporal consistency across frames. To address this, we propose the Space-Time Depth Estimation. To more accurately comprehend the spatial motion information in panoramic videos, we employ a panoramic optical flow estimation model to derive the optical flow $\mathbf{F}\in \mathbb{R}^{(L-1) \times H \times W \times 2}$ of the panoramic video, representing pixel movement between consecutive frames. 
We define the mask $ \mathbf{M}_l \in \mathbb{R}^{H \times W}$ to highlight pixels with motion distances, where $ l = 1, \ldots, L-1 $. We then compute the post-motion positions of these pixels to obtain $ \mathbf{M}^{\prime}_l \in \mathbb{R}^{H \times W}$, where $ l = 2, \ldots, L $. We use the following formula to calculate the motion region $\mathbf{M}^l$ for each frame,  identifying areas that have changed since the previous frame or will generate new motion in the next one:
\begin{equation}
\mathbf{M}^l=\mathbf{M}_{l-1} \vee \mathbf{M}^{\prime}_{l} \vee \mathbf{M}_{l} ,\quad\tilde{\mathbf{M}}=\bigvee_{l=1}^{L} \mathbf{M}^l
\label{eq:mask}
\end{equation}
The overall motion region mask $\tilde{\mathbf{M}}$ is the union of the masks for each frame. 

For subsequent panoramic frames $(l > 1)$, We perform the space-time depth estimation to ensure spatial and temporal consistency in the depth maps. This process leverages motion regions from optical flow for adaptive perspective selection. Additionally, the previously estimated depth maps offer supervision. The goal of optimization is extended for space-time depth alignment as follows:
\begin{align}\label{eq:st-align}
\min \operatornamewithlimits{\mathbb{E}}_{i \in S} \{&\mathcal{L}_{\text{spatial}}\} + 
\lambda_{\text{first}}\mathcal{L}_{\text{first}} +
\lambda_{\text{pre}}\mathcal{L}_{\text{pre}} \\ \nonumber
\text{where} \quad &\mathcal{L}_{\text{spatial}} = \lambda_{\text{depth}} \mathcal{L}_{\text{depth}}  + 
\lambda_{\text{scale}}\mathcal{L}_{\text{scale}} +
\lambda_{\text{shift}}\mathcal{L}_{\text{shift}} \\ \nonumber
&\mathcal{L}_{\text{first}} = \|(\mathbf{D}^1 - \operatorname{MLPs}(\mathbf{v}_\text{pano}; \boldsymbol{\Theta}^{l})) \circ \neg{\tilde{\mathbf{M}}}\| \\ \nonumber
&\mathcal{L}_{\text{pre}} = \|(\mathbf{D}^{l-1} - \operatorname{MLPs}(\mathbf{v}_\text{pano}; \boldsymbol{\Theta}^{l})) \circ \tilde{\mathbf{M}} \circ \neg \mathbf{M}^l\| \nonumber
\end{align}
where $S = \{ i \mid \exists j, k \text{ such that } \gamma(\mathbf{M}^l, \mathbf{v}_i)[j, k] = \text{True} \}$, and $\gamma$ is the projection function that maps a panoramic image to a perspective image based on direction $\mathbf{v}$. Consequently, only the perspective images that overlap with the motion region mask $\mathbf{M}^l$ are involved for optimization of the current frame $\mathbf{f}^l$. Other areas of the overall motion region $\tilde{\mathbf{M}}$ are supervised using depth $\mathbf{D}^{l-1}$ from the previous frame $\mathbf{f}^{l-1}$, while non-motion regions are supervised using depth $\mathbf{D}^{1}$ from the first frame $\mathbf{f}^1$. Using Eq. \ref{eq:st-align}, we could estimate the panoramic depth map $\mathbf{D}^l (l>1)$ frame by frame. thereby obtaining the depth of the entire panoramic video.

\subsubsection{4D Scene Reconstruction}\label{sec:representaion}
After completing the depth estimation, the panoramic video and its depth map $\{[\mathbf{f}^l, \mathbf{D}^l]\}, (l=1,2,...L)$ are converted into a 4D point cloud with temporal attributes, which serves as the initialization for the 4D scene. We choose Spacetime Gaussian \cite{spacetimegaussian} as the 4D representation of the scene. The video is projected into perspective views with varying FoV for supervision during training. Since the camera position of the panoramic video is fixed, we project the panoramic depth into the corresponding perspective depth map and apply depth-based warping to generate new views by perturbing the camera position relative to the original view. This process enriches the training set and improves scene completeness and rendering robustness.

\subsection{360World Dataset}\label{sec:dataset}
Current large-scale text-video datasets like WebVid \cite{webvid} primarily feature narrow-FoV perspective videos instead of panoramic videos. Furthermore, existing datasets \cite{360dvd,imagine360} for panoramic video generation primarily include footage captured with moving cameras, which makes them unsuitable for the task of 4D scene generation.
%capturing scenes in equirectangular projection with large-scale spatial motion, exhibiting non-rigid flow effects, which can complicate the model's spatial understanding of the scenes.
To address data limitations, we present the \textit{360World dataset}, consisting of $7,497$ high-quality panoramic video clips with a total of $5,380,909$ frames. Each clip is accompanied by textual descriptions sourced from open-domain content. The videos capture a wide array of real-world scenarios, from natural landscapes to urban environments, offering robust data support for generative models to comprehend dynamic panoramic scenes.

We collect original YouTube videos and annotate the segmented clips. we employ ShareGPT4Video \cite{sharegpt4video}, a Large Video-Language Model (LVLM)  with strong video comprehension capabilities, to deeply analyze the video across both spatial and temporal dimensions and generate detailed text prompts of panoramic videos. Finally, we utilize a Large Language Model (LLM) for post-processing the text, summarizing and refining the detailed prompt by removing photography-related descriptive terms such as ``camera'' and ``video'', resulting in final text prompts that effectively describe the scene content and dynamic motions.
% Enunciations

\begin{figure*}
  \includegraphics[width=\textwidth]{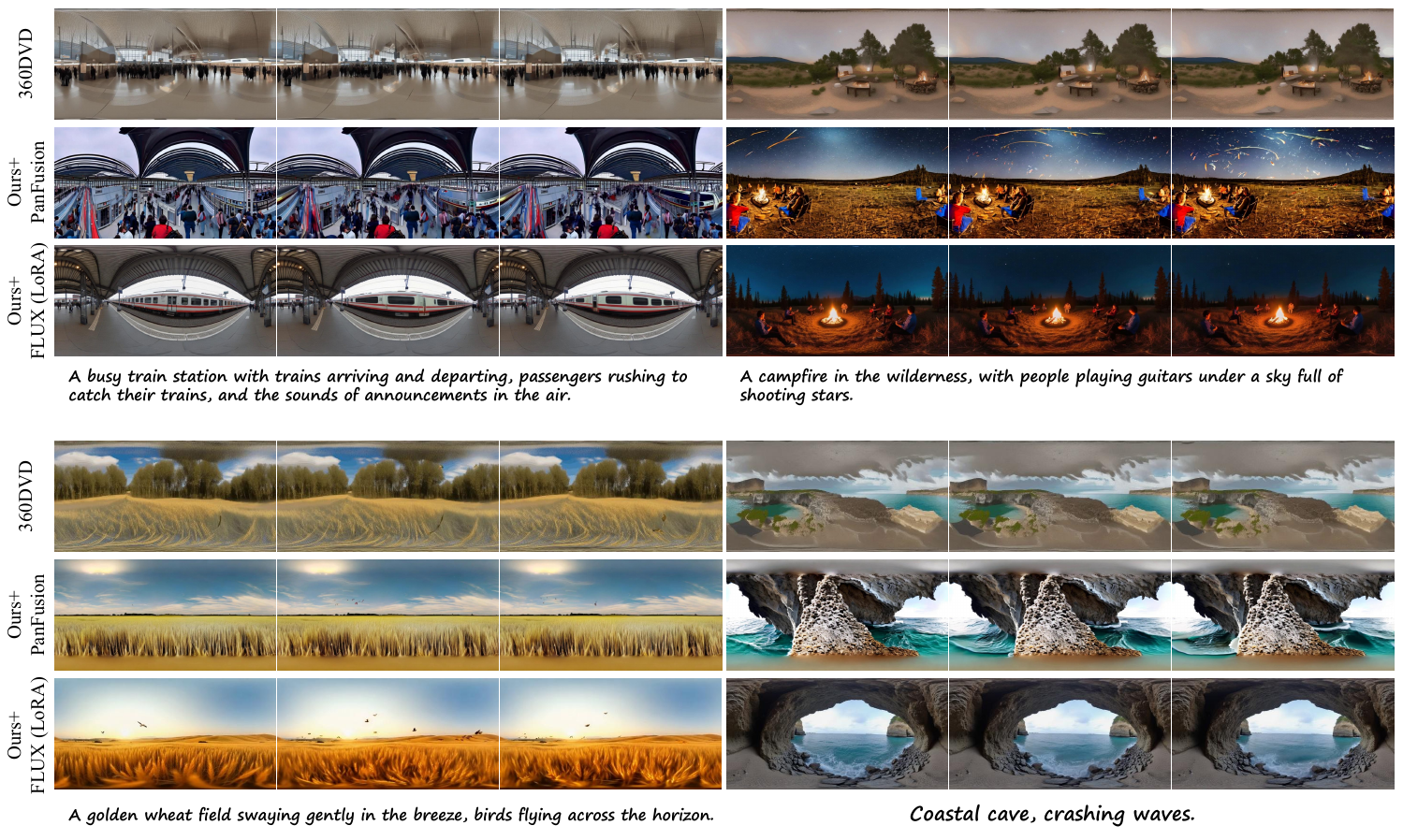}
  \caption{Qualitative comparison of text-driven panoramic video generation. Our Panoramic Animator effectively achieves more coherent motion and avoids the occurrence of artifacts.}
  \label{fig:video_com}
\end{figure*}

% Start of "Sample References" section

\section{Experiments}
In this section, we first introduce the implementation details in Sec. \ref{sec:detail}, followed by a multi-level and comprehensive evaluation. We compare our \textit{Panoramic Animator} with other panoramic video generation methods in Sec. \ref{sec:videocomp}. We then evaluate \textit{HoloTime} against other image-driven 4D scene generation methods in Sec. \ref{sec:scenecomp}. Finally, we conduct ablation experiments in Sec. \ref{sec:ablation} to validate the effectiveness of proposed techniques.

\subsection{Implementation Details}\label{sec:detail}
In the implementation of \textit{Panoramic Animator}, the frame length $L$ of the video is set to $25$. The resolution of the guidance model is $320\times512$. We only train the spatial layers from pre-trained weights for $5,000$ iterations on the \textit{360World dataset}. For the refinement model, a progressive training strategy from ViewCrafter \cite{viewcrafter} is applied. Initially,the entire UNet is trained on the hybrid data at $320\times512$ resolution for $5,000$ iterations, then the spatial layers are further trained at $576\times1024$ resolution on the \textit{360World dataset} for another $5,000$ iterations to upscale the resolution. Throughout the training, the batch size is set to $16$, and the learning rate to $5\times10^{-5}$. We employ Real-ESRGAN\cite{realesrgan} as the super-resolution model. The final generated panoramic video frames are resized to $512\times1024$ to maintain the panoramic aspect ratio.

During the reconstruction phase, the generated panoramic video is first upscaled to $1024\times2048$ using Real-ESRGAN. The number of perspective views $L$ is set to $20$, with each perspective image having a resolution of $512\times512$. Depth estimation for these images is conducted with Marigold \cite{marigold}. PanoFlow \cite{panoflow} is utilized for optical flow estimation of the panoramic videos. We train the Spacetime Gaussian \cite{spacetimegaussian} lite model for
30,000 iterations with the default hyperparameter settings for each scene.

\begin{table}%
\setlength{\tabcolsep}{2.5pt}
\caption{User study of text-driven panoramic video generation. Our approach can integrate
different personalized text-to-panorama generation models with outperformance results.}
\label{tab:text_video_com}
\begin{center}
\scalebox{0.9}{
\begin{tabular}{l|cc|ccc}
  \toprule
  \multirow{2}*{Method} &
                \multicolumn{2}{c}{Video Criteria} &
                \multicolumn{3}{c}{Panorama Criteria} \\
            \cline{2-6}
  ~ & \thead{Graphics\\Quality} & \thead{Frame\\Consistency} & \thead{End\\Continuity} & \thead{Content\\Distribution} &
        \thead{Motion\\Pattern} \\ \midrule
  360DVD \shortcite{360dvd} & 5.6692 & 5.9462 & 6.1308 & 5.7731 & 5.6385\\
  Ours+PanFusion \shortcite{panfusion} & 6.0962 & 6.3615 & 6.5308 & 5.9731 & 6.1692\\
  Ours+FLUX (LoRA) & \textbf{8.1423} & \textbf{8.1538} & \textbf{8.3500} & \textbf{8.2000} & \textbf{8.0308}\\
  \bottomrule
\end{tabular}
}
\end{center}
\bigskip\centering
\end{table}%

\begin{figure*}
  \includegraphics[width=\textwidth]{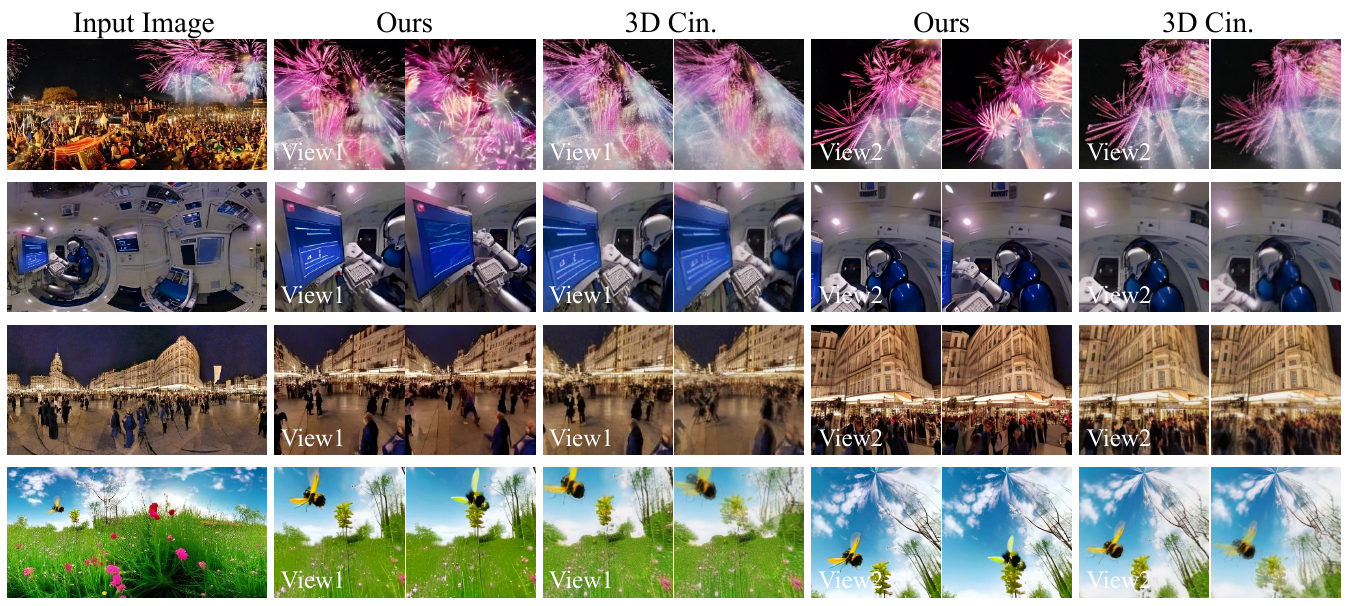}
  \caption{Qualitative comparison of image-driven 4D scene generation. Our method can generate more complex and diverse motions in the scene while maintaining spatial and temporal consistency in the dynamic scene.}
  \label{fig:scene_com}
\end{figure*}

\begin{table*}%
\caption{Quantitative comparison of image-driven panoramic video generation. 360DVD* is fine-tuned on DynamiCrafter using the techniques proposed by 360DVD  with the 360World dataset. “pers” indicates that the panoramic videos are projected into perspective videos for evaluation. Our method outperforms the baseline in terms of all metrics.}
\label{tab:image_video_com}
\begin{center}
\scalebox{1}{
\begin{tabular}{l|ccccc|cc}
  \toprule
  \multirow{2}*{Method} &
                \multicolumn{5}{c}{VBench (pers)} &
                \multicolumn{2}{c}{ChronoMagic-Bench} \\
            \cline{2-8}
		~ & \footnotesize Subject Consistency & \footnotesize Background Consistency & \footnotesize Temporal Flickering & \footnotesize Motion Smoothness & \footnotesize Dynamic Degree & 
        \footnotesize GPT4o MTScore & \footnotesize 
        MTScore\\ \midrule
  360DVD* \shortcite{360dvd} & 0.9508 & 0.9489 & 0.9797 & 0.9843 & 0.0667 & 2.2000 & 0.3556 \\
  Ours & \textbf{0.9543} & \textbf{0.9501} & \textbf{0.9864} & \textbf{0.9903} & \textbf{0.0817} & \textbf{2.4111} & \textbf{0.3629}\\
  \bottomrule
\end{tabular}
}
\end{center}
\bigskip\centering
\end{table*}%

\subsection{Panoramic Video Generation Comparisons}\label{sec:videocomp}
Due to the absence of available image-driven panoramic video generation methods currently, we compare our \textit{Panoramic Animator} with 360DVD \cite{360dvd}, a text-driven panoramic video generation method fine-tuned on AnimateDiff \cite{animatediff}. Our approach can integrate different text-to-panorama generation models to achieve text-driven panoramic video generation. PanFusion \cite{panfusion} and FLUX \cite{flux} with its Panorama LoRA \cite{lora} from Hugging Face are employed to conduct comparison. Given the input text prompt, They first creates a panoramic image, which is then used by the \textit{Panoramic Animator} to generate the panoramic video.

We use a Large Language Model (LLM) to generate a batch of text prompts for various scenes, emulating the captions from the \textit{360World dataset}, and use the generated prompts as the input for text-driven comparison. Fig. \ref{fig:video_com} shows the results of qualitative comparison, highlighting broad applicability of Panoramic Animator. We conduct a user study to comprehensively evaluate generated panoramic videos in terms of visual criteria and panoramic criteria. We follow the criteria set in 360DVD, including graphics quality, cross-frame consistency, left-right continuity, content distribution, and motion patterns. $26$ participants rate each metric of the $10$ sets of generated videos on a scale from $1$ to $10$. We calculate the average score of the different metrics. Table. \ref{tab:text_video_com} presents the results of user study, revealing that our method not only achieve high video quality but also effectively align with the characteristics of panoramic videos. This also demonstrates the strong adaptability of our method for multiple personalized text-to-panorama models.

To conduct a more precise and targeted comparison, we employ the techniques proposed by 360DVD to fine-tune the same base model DynamiCrafter with our \textit{360World dataset}, thereby obtaining 360DVD*, for the purpose of image-driven panoramic video generation comparison. We use multiple panoramic image generation models \cite{panfusion,flux,holodreamer} to generate $90$ panoramic images of different styles based on the generated prompts as input. Panoramic videos are projected into perspective videos to compute the metrics of VBench \cite{vbench} for evaluating video details, including Subject Consistency, Background Consistency, Temporal Flickering, Motion Smoothness and Dynamic Degree. The MTScore (Metamorphic Score) metrics of ChronoMagic-Bench \cite{chronomagic} are used to directly assess the global motion of panoramic videos. Table. \ref{tab:image_video_com} demonstrates that our proposed techniques can achieve better temporal and motion details. Higher GPT4o MTScore and MTScore indicates that our method can generate more significant overall motion amplitude.
%We used 60 text prompts and their corresponding generated videos for evaluation. 

\begin{table}%
\setlength{\tabcolsep}{2.5pt}
\caption{Quantitative comparison and user study of image-driven 4D scene generation. Our method outperforms the baselines in terms of all metrics.}
\label{tab:scene_com}
\begin{center}
\scalebox{0.85}{
\begin{tabular}{l|cccc|cc}
  \toprule
  \multirow{2}*{Method} &
                \multicolumn{4}{c}{Q-Align} &
                \multicolumn{2}{c}{User Study} \\
            \cline{2-7}
  ~ & \thead{Image\\Quality} & \thead{Image\\Aesthetics} & \thead{Video\\Quality} & \thead{Video\\Aesthetics} & \thead{Graphics\\Quality} & \thead{Temporal\\Coherence} \\ \midrule
  3D Cin. (zoom in) & 1.9175 & 1.6271 & 2.266 & 1.8446 & 10.32\% & 11.29\%\\
  3D Cin. (circle) \shortcite{3dcind}& 1.9433 & 1.6150 & 2.2823 & 1.8365  & 1.94\% & 5.16\%\\
  Ours & \textbf{2.2643} & \textbf{1.7627} & \textbf{2.6293} & \textbf{2.0350} & \textbf{87.74\%} & \textbf{83.55\%}\\
  \bottomrule
\end{tabular}
}
\end{center}
\end{table}%

\subsection{4D Scene Generation Comparisons}\label{sec:scenecomp}We compare our framework with the optical flow-based 3D dynamic image technology, 3D-Cinemagraphy (3D-Cin.) \cite{3dcind}. Following the experimental setup of 4K4DGen, we employ 3D-Cin. to construct 4D scenes from the input panoramic images under the "circle" and "zoom-in" settings, and then project the rendered videos into perspective videos for comparison. Fig. \ref{fig:scene_com} shows the results of the qualitative comparison. It is evident that the optical flow-based method is mainly effective for creating flowing effects, making it suitable primarily for fluids like water, which limits its applications. In contrast, our method utilizes a video diffusion model, generating more complex texture variations and spatial motion, and thus exhibits superior generalization capabilities.

We employ the Q-Align \cite{qalign} metrics to evaluate the quality and aesthetic scores of the perspective videos rendered from the generated scenes, as well as the individual video frames. We also conduct a 4D scene generation user study, where 31 participants evaluate 10 groups of generated 4D scenes and select the best method for each group of scenes based on the criteria of graphics quality and temporal coherence. Our method achieves better scores on all metrics in Table. \ref{tab:scene_com}, highlighting the quality of the generated scenes.

\begin{table}[t]%
\caption{Quantitative ablation study of hybrid data fine-tuning (HDF) and motion guided generation (MGG). “pers” indicates that the panoramic videos are projected into perspective videos for evaluation.}
\label{tab:video_ab}
\begin{center}
\scalebox{0.95}{
\begin{tabular}{c|ccc|cc}
  \toprule
  \multirow{2}*{Method} &
                \multicolumn{3}{c}{VBench (pers)} &
                \multicolumn{2}{c}{\small ChronoMagic-Bench} \\
            \cline{2-6}
		~ & \thead{Temporal\\Flickering} & 
        \thead{Motion\\Smoothness} & \thead{Dynamic\\Degree} &
        \thead{CHScore} & 
        \thead{GPT4o\\MTScore} \\ \midrule
  w/o HDF  & 0.9860 & 0.9898 & 0.0694 & 533.8 & \textbf{2.4556} \\
  w/o MGG  & 0.9861 & 0.9899 & 0.0794 & 560.4 & 2.1444 \\
  Ours Full & \textbf{0.9864} & \textbf{0.9903} & \textbf{0.0816}  & \textbf{647.9} & 2.4111 \\
  \bottomrule
\end{tabular}
}
\end{center}
\end{table}%

\begin{figure}[tp]
  \includegraphics[width=\columnwidth]{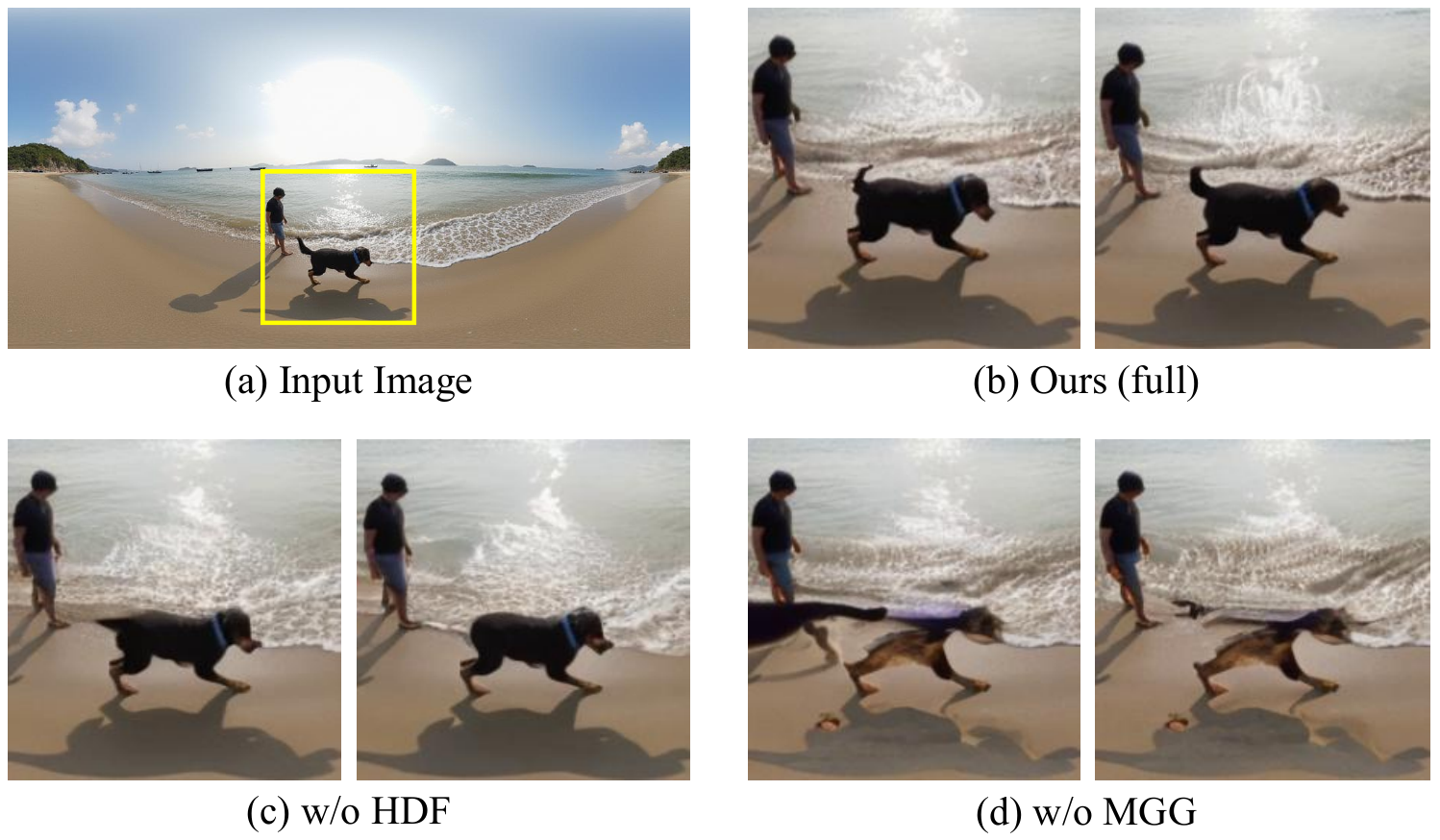}
  \caption{Ablation study of hybrid data fine-tuning (HDF) and motion guided generation (MGG) for panoramic video generation.}
  \label{fig:video_ab}
\end{figure}

\subsection{Ablation Study}\label{sec:ablation}
\begin{figure}
  \includegraphics[width=\columnwidth]{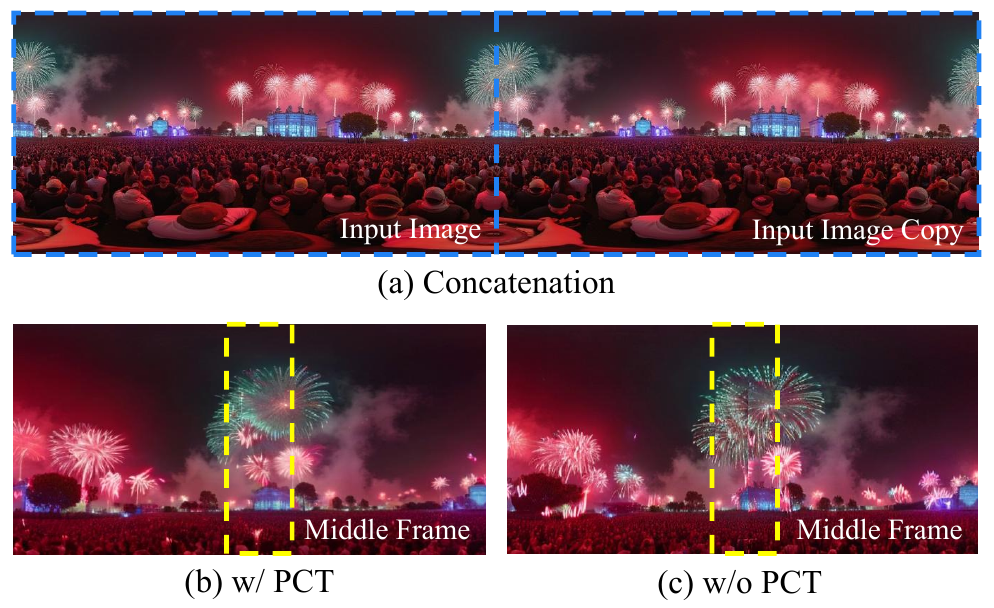}
  \caption{Ablation study of panoramic circular techniques (PCT). We concatenate the left and right ends of the generated frames to check for continuity. PCT effectively prevents the occurrence of discontinuous seam.}
  \label{fig:video_circular}
\end{figure}

We perform ablation experiments on \textit{Panoramic Animator} and \textit{Panoramic Space-Time Reconstruction} separately. Firstly, we evaluate the impact of our proposed Hybrid Data Fine-tuning (HDF) and Two-Stage Motion Guided Generation (MGG) on the \textit{Panoramic Animator}, as illustrated in Fig. \ref{fig:video_ab}. We perform quantitative assessments of HDF and MGG in Table. \ref{tab:video_ab}. We employ three temporal metrics from VBench to assess projected perspective videos, as well as CHScore (Coherence Score) and GPT4o MTScore from ChronoMagic-Bench for panoramic videos. The results illustrate the contribution of HDF to enhancing temporal details and coherence, as well as the impact of MGG on the overall motion. Meanwhile, Fig. \ref{fig:video_circular} demonstrates the effectiveness of PCT, which prevents the occurrence of discontinuous seam.

\begin{figure}
  \includegraphics[width=\columnwidth]{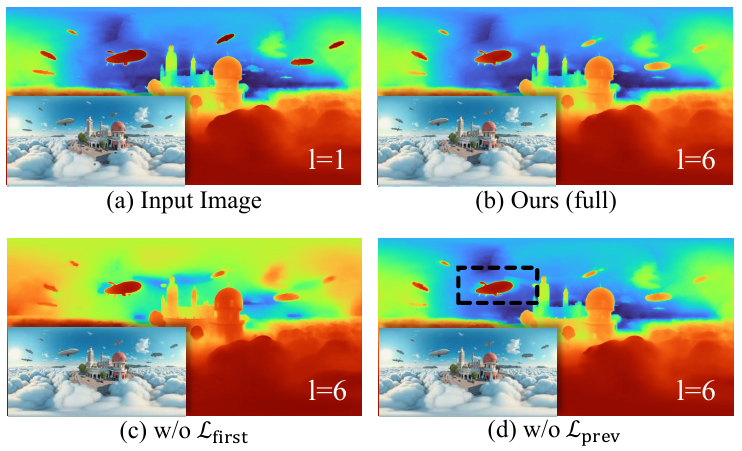}
  \caption{Ablation study of temporal loss terms for Space-Time Depth Estimation. Based on the same first frame depth, our temporal loss terms contribute to higher temporal consistency of the middle frame's depth.}
  \label{fig:depth_ab}
\end{figure}

We also assess the effectiveness of the temporal loss terms in Eq. \ref{eq:st-align} for \textit{Space-Time Depth Estimation}. As illustrated in Fig. \ref{fig:depth_ab}, for panoramic videos with significant spatial motion, the $\lambda_{\text{first}}$ term ensures overall depth consistency between frames, while the $\lambda_{\text{prev}}$ term mitigates artifacts in regions near motion.

\section{Conclusions}
In this work, we present \textit{HoloTime}, an novel framework that transforms static panoramic images into large-scale 4D scenes. To tackle the limited availability of panoramic video datasets, we introduce \textit{360World}, a comprehensive dataset of fixed-camera panoramic videos. We propose the \textit{Panoramic Animator}, which utilizes an I2V diffusion model for direct panoramic video generation. Moreover, we introduce \textit{Panoramic Space-Time Reconstruction}, a method for 4D reconstruction of panoramic videos that ensures temporal and spatial consistency. Our approach outperforms existing methods by creating more captivating and realistic immersive dynamic environments, enhancing the virtual roaming experience.

\bibliographystyle{ACM-Reference-Format}
\bibliography{sample-base}

\appendix

\section{360World Dataset Processing}

We only download publicly available YouTube videos for training. We utilize keywords such as ``panorama'' and ``VR'' for retrieval and filtering. Given that each original video may contain multiple distinct scenes and transitions between them, we perform keyframe-based slicing on the original videos to identify frames with significant changes. Due to the camera's broad field of view (FoV), there are typically not large areas of pixel changes throughout the real-world panoramic video. Most motion is confined to specific local regions within the panoramic space. Therefore, we use the pixel distance between adjacent frames to determine whether a frame is a keyframe. 
Firstly, we perform pre-processing on each video. To facilitate the detection of significant changes between frames, we sample video frames at a fixed interval along the temporal dimension, thereby increasing the video speed. 

For a sampled video with \(L\) frames, each frame is converted into a grayscale image. The \(i\)-th grayscale image is defined as $g_{i} \in \mathbb{R}^{H \times W}$. We calculate the pixel distance between current frame $g_{i}$ and the previous frame $g_{i-1}$. If the distance at any pixel exceeds the threshold $\theta_{\text{trans}}$, that pixel is labeled a transition pixel. If the proportion of transition pixels across the entire frame surpasses the threshold $\theta_{\text{count}}$ across the entire frame, it signifies a significant change area, classifying it as a keyframe. The formal formula is as follows:

\begin{equation}
f(D, \theta_1) = \left| \{ (i, j) \mid d_{ij} > \theta_1, 1 \leq i \leq h, 1 \leq j \leq w \} \right|
 \label{eq:count}
\end{equation}

\begin{equation}
\mathcal{T}_i = \begin{cases} 
\text{True,} & \text{if } f(\left|g_i-g_{i-1}\right|, \theta_{\text{trans}}) > \theta_{\text{count}} \\
\text{False,} & \text{otherwise}
\end{cases}
 \label{eq:keyframe}
\end{equation}

After detecting keyframes, we slice the original video and discard frames that are too close to the keyframes to eliminate complex transitions involving multiple frames. 

\begin{figure}
  \includegraphics[width=\columnwidth]{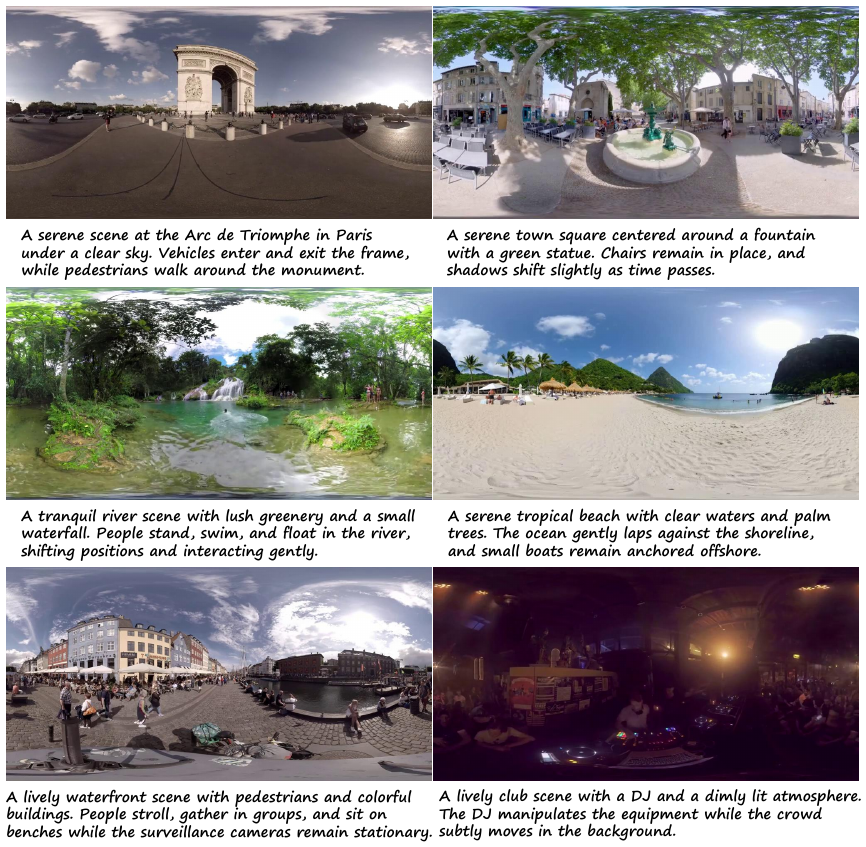}
  \caption{Samples in the 360World dataset.}
  \label{fig:video_com}
\end{figure}

\section{Implementation Details}
\subsection{Panoramic Animator}

The training of Panoramic Animator is completed on 8 NVIDIA A100 GPUs with 80 GB RAM. For inference, we employ the DDIM sampler with classifier-free guidance. Initially, we copy the left one-fifteenth of the reference image and concatenate it to the right end. The timesteps $T$ is set to $1,000$, while $K$ is set to $800$. After each denoising step, the left and right one-sixteenths are blended, as shown in Panoramic Circular Techniques (PCT). Finally, we crop the right one-sixteenth of the generated panoramic video.

\subsection{Panoramic Space-Time Reconstruction}
In the phase of depth estimation, following DreamScene360 \cite{dreamscene360}, we project each panorama frame into $20$ perspective views based on the surface directions of a regular icosahedron. The hyperparameter settings of depth optimization are as follows: $\lambda_{\text{depth}}=\lambda_{\text{shift}}=\lambda_{\text{first}}=\lambda_{\text{pre}}=1$, $\lambda_{\text{scale}}=0.1$. When optimizing the spatial depth alignment of the first frame, we perform $3,000$ iterations, with $\lambda_{\text{shift}}=0$ for the first $1,500$ iterations. Subsequent frame depth alignments consist of $1,000$ iterations each. continuously optimizing the current geometric field $\boldsymbol{\Theta}^{l}$ based on the optimization parameters $\boldsymbol{\Theta}^{l-1}$ from the proir.s

In the training of Spacetime Gaussians \cite{spacetimegaussian}, We set up $38$ cameras with approximately $45°$ FoV (Field of View) to cover the spherical perspective of the panoramic video. The depth-based warping is applied to generate $4$ supplementary views for each view, with camera position disturbances of $0.1$ in the up, down, left, and right directions. Additionally, $20$ cameras with approximately $75°$ FoV are set up. For each scene, We train the lite model of Spacetime Gaussians for $30,000$ iterations on single NVIDIA A800 GPU with 80 GB RAM, using the default hyperparameter settings.

\subsection{4D Point Cloud Initialization}
To reduce memory usage and improve training efficiency of Spacetime Gaussianss, we decrease the initialized 4D point cloud. A 4D point cloud used for Spacetime Gaussian initialization has a time attribute $t$, which indicates the moment the point exists. First, we convert the panoramic video into a grayscale image and calculate the standard deviation of each pixel over time. For pixels with a standard deviation less than $20$, we mark them as texture variation areas $\mathbf{M}_\text{std}$. For the first frame of the video at time $t=0$, all pixel points are merged into the 4D point cloud. For subsequent frames at time $t$, only points that covered by $\mathbf{M}_\text{std}$ or the optical flow detected motion area $\mathbf{M}^l$ of the current frame are merged into the 4D point cloud.

\section{Experiments Details}
We use $90$ prompts their generated panoramic images to evaluate panoramic video generation, as well as $20$ generated panoramic images for assessing 4D scene generation. We employ a projection method based on a regular icosahedron in experiments to project panoramic videos to render perspective videos. All of the panoramic images used in experiments and presentations are generated by PanFusion \cite{panfusion}, FLUX \cite{flux} (panorama LoRA), and panorama generation method in Holodreamer \cite{holodreamer}. When designing these experimental data, in addition to the realistic style, we assigned some other styles such as cartoon, cyberpunk, etc., to a portion of the prompts and images, in order to evaluate the out-of-distribution (OOD) generalization capability of our method.

We replace the original 360DVD base model with DynamiCrafter \cite{dynamicrafter} and complete 15,000 iterations of training using our 360World dataset under the same hyperparameter settings to obtain 360DVD* for comparison in image-driven panorama generation.

\section{Visual Results}
We provide a supplementary video containing the generated results,
and we strongly recommend that you view it to have the most
intuitive understanding of the generated 4D scenes and panoramic
videos. We use Venhancer \cite{venhancer} to super-resolution the generated
panoramic videos before Panoramic Space-Time Reconstruction
for a better VR immersive experience of 4D scenes. This technique is applied solely to the 4D scene presentations in the supplementary video and is not used elsewhere in the video. Notably, it is also excluded from any comparative experiments or ablation studies in the main paper.

\end{document}